\relax
\documentclass[letterpaper]{article} 
\usepackage{aaai19}  
\usepackage{times}  
\usepackage{helvet}  
\usepackage{courier}  
\usepackage{url}  
\usepackage{graphicx}  
\usepackage{amsmath}
\usepackage{float}
\usepackage{amsthm}
\usepackage{subfig}
\usepackage{amssymb}
\newtheorem{theorem}{Theorem}
\newtheorem*{remark}{Remark}
\begin{document}
%
\title{Complex Unitary Recurrent Neural Networks using Scaled Cayley Transform}

\author{Kehelwala D. G. Maduranga, Kyle E. Helfrich, and Qiang Ye\\
 Mathematics Department, University of Kentucky\\ Lexington, KY, 40508, United States\\
 \{kdgmaduranga,kyle.helfrich,qye3\}@uky.edu\\
 }

\maketitle

\begin{abstract}

Recurrent neural networks (RNNs) have been successfully used on a wide range of sequential data problems. A well known difficulty in using RNNs is the \textit{vanishing or exploding gradient} problem.  Recently, there have been several different RNN architectures that try to mitigate this issue by maintaining an orthogonal or unitary recurrent weight matrix.  One such architecture is the scaled Cayley orthogonal recurrent neural network (scoRNN) which parameterizes the orthogonal recurrent weight matrix through a scaled Cayley transform. This parametrization contains a diagonal scaling matrix consisting of positive or negative one entries that can not be optimized by gradient descent. Thus the scaling matrix is fixed before training and a hyperparameter is introduced to tune the matrix for each particular task.  In this paper, we develop a unitary RNN architecture based on a complex scaled Cayley transform.  Unlike the real orthogonal case, the transformation uses a diagonal scaling  matrix consisting of entries on the complex unit circle which can be optimized using gradient descent and no longer requires the tuning of a hyperparameter. We also provide an analysis of a potential issue of the modReLU activiation function which is used in our work and several other unitary RNNs. In the experiments conducted, the scaled Cayley unitary recurrent neural network (scuRNN) achieves comparable or better results than scoRNN and other unitary RNNs without fixing the scaling matrix. 
\end{abstract}

\section{Introduction}

Recurrent neural networks (RNNs) have been successfully used on a wide range of sequential data problems. A main difficulty when training RNNs using a gradient descent based optimizer is the \textit{vanishing or exploding gradient} problem \cite{Bengio94}. The exploding gradient problem refers to the large growth of gradients as they propagate backwards through time and the vanishing gradient problem occurs when the gradients tend toward zero. The exploding gradient case will cause the trainable parameters to vary drastically during training, resulting in unstable performance. For the vanishing gradient case, training will progress slowly, if at all.

A range of different architectures have been proposed to address this problem. Currently, the most common architectures involve gating mechanisms that control when information is retained or discarded such as the Long Short-Term Memory networks (LSTMs) \cite{Hochreiter97} and Gated Recurrent Units (GRUs) \cite{ChoMGBSB14}.  More recently, several architectures have been proposed that maintain a unitary recurrent weight matrix. The unitary evolution RNN (uRNN) architecture proposed by \cite{Arjo16} maintains a unitary matrix by using a product of simple unitary matrices. The full-capacity uRNN \cite{Wisdom16} maintains a general unitary matrix by optimizing along a gradient descent direction on the Stiefel manifold. The tunable efficient unitary neural network (EURNN) by \cite{jing16} constructs the unitary matrix using a product of complex rotation matrices. For additional work with unitary RNNs and complex valued networks, see \cite{Hyland17}, \cite{jing17}, \cite{vorontsov17}, \cite{Wolter17} and \cite{trabelsi18}.  There have also been several architectures that maintain an orthogonal recurrent weight matrix. The orthogonal RNN (oRNN) by \cite{Mhammadi16} uses a product of Householder reflection matrices while the  scaled Cayley orthogonal RNN (scoRNN) architecture parameterizes the recurrent weight matrix by a skew-symmetric matrix and a diagonal matrix through the Cayley transform \cite{kyle17}. Compared with other unitary/orthogonal RNNs, the scoRNN architecture has a simple parameterization that has been shown to be advantageous in \cite{kyle17}.  The exploding or vanishing gradient problem has also been examined in \cite{Henaff17} and \cite{Le15}. 

In this paper, we address a difficulty of scoRNN. The scoRNN parameterization of the orthogonal recurrent weight matrix contains a diagonal matrix consisting of $\pm 1$ on the diagonal. These discrete parameters are used to define the scaling matrix, which may critically affect performance, and can not be optimized by gradient descent. Thus scoRNN introduces a tunable hyperparameter that controls the number of negative ones on the diagonal. This hyperparameter is tuned for each particular task for optimal results.  This causes the scaling matrix to remain fixed during training. We propose a method to overcome this difficulty by using a unitary recurrent weight matrix parameterized by a skew-Hermitian matrix and a diagonal scaling matrix through the scaled Cayley transform, where the entries of the diagonal matrix lie on the complex unit circle and have the form $e^{i \theta}$. This parameterization is differentiable with respect to the continuous $\theta$ variable and can be optimized using gradient descent. This eliminates the need for tuning a hyperparameter and having a fixed scaling matrix during training. We call this new architecture the scaled Cayley unitary recurrent neural network (scuRNN). We also develop the update scheme to train the skew-Hermitian and diagonal scaling matrices. The experiments performed show that scuRNN achieves better or comparable results than other unitary RNNs and scoRNN without the need for tuning an additional hyperparameter.

For many unitary RNNs, a popular activation function is the modReLU function \cite{Arjo16}. Known architectures that incorporate the modReLU function include works by \cite{Arjo16}, \cite{Wisdom16}, \cite{jing16}, \cite{jing17}, \cite{kyle17}, and \cite{Wolter17}. We also use the modReLU activation function in this work but have noticed a singularity issue that may potentially impact performance. To the best of our knowledge, this singularity has not been previously discussed in the literature. In section \ref{modReLU}, we provide an analysis of the modReLU function and discuss initialization schemes that may mitigate the singularity. 

We note that there has been recent interest in complex networks outside of uRNN as discussed in the papers by \cite{trabelsi18} and \cite{Wolter17}. Our work presents an additional case where complex networks can be advantageous over real networks. 

\section{Background}

\subsection{Real RNNs}

\noindent A single hidden layer recurrent neural network (RNN) is a dynamical system that uses an input sequence $\textbf{x} = (\textbf{x}_1, \textbf{x}_2, ... , \textbf{x}_{\tau})$ where each $\textbf{x}_i \in \mathbb{R}^m$, to produce an output sequence $\textbf{y} = (\textbf{y}_1, \textbf{y}_2, ... , \textbf{y}_{\tau})$ with $\textbf{y}_i \in \mathbb{R}^p$ given recursively by the following: 
\begin{equation}
\label{eq:1}
\textbf{h}_t  = \sigma \left(U\textbf{x}_t + W\textbf{h}_{t-1} + \textbf{b}\right) \; ; \;\textbf{y}_t  = V\textbf{h}_t + \textbf{c}
\end{equation}

\noindent where $U \in \mathbb{R}^{n \times m}$ is the input to hidden weight matrix, $W \in \mathbb{R}^{n \times n}$ the recurrent weight matrix, $\mathbf{b} \in \mathbb{R}^n$ the hidden bias, $V \in \mathbb{R}^{p \times n}$ the hidden to output weight matrix, and $\mathbf{c} \in \mathbb{R}^p$ the output bias.  Here $m$ is the input data size, $n$ is the number of hidden units, and $p$ is the output data size. The sequence $\mathbf{h} = (\mathbf{h}_0, \ldots, \mathbf{h}_{\tau-1})$, is the sequence of hidden layer states with $\mathbf{h}_i \in \mathbb{R}^n$ and $\sigma(\cdot)$ is a pointwise nonlinear activation function, such as a hyperbolic tangent function or rectified linear unit \cite{Nair10}.

For long sequence lengths, RNNs are prone to suffer from the exploding or vanishing gradient problem. As detailed in \cite{Arjo16}, the exploding gradient problem can occur when the spectral radius of the recurrent weight matrix is greater than one and the vanishing gradient problem can occur when it is less than one. Maintaining a strict unitary or orthogonal recurrent weight matrix with a spectral radius of one can help mitigate this problem. 

\subsection{Unitary RNNs}

Similar to orthogonal matrices, unitary matrices are complex matrices $W \in \mathbb{C}^{n \times n}$ with spectral radius one and the property that $W^{*}W = I$, where $*$ denotes the conjugate transpose operator and $I$ is the identity matrix. All unitary RNNs are designed to maintain a unitary recurrent weight matrix. In this section, we examine the unitary evolution RNN (uRNN), full-capacity uRNN, and the EURNN. For notational purposes, we follow the convention established in \cite{Wisdom16} and refer to the unitary evolution RNN as the restricted-capacity uRNN. 

The restricted-capacity uRNN maintains a unitary recurrent weight matrix by using a parameterization consisting of diagonal matrices with entries lying on the complex unit disk, Householder reflection matrices, Fourier and inverse Fourier transform matrices, and a fixed permutation matrix \cite{Arjo16}. As shown by \cite{Wisdom16}, this parameterization contains only $7n$ trainable parameters and is unable to represent all unitary matrices when the hidden size $n > 7$. 

The full-capacity uRNN does not parameterize the recurrent weight matrix directly, but restricts the descent direction to the Stiefel manifold $\{W \in \mathbb{C}^{n \times n} | W^{*}W = I \}$. This is done by traveling along a curve of the tangent plane projected onto the Stiefel manifold using a multiplicative update scheme as outlined in \cite{Wisdom16}.  As shown in \cite{kyle17}, this multiplicative update scheme may result in a loss of orthogonality due to numerical rounding issues. 

Similar to the restricted-capacity uRNN, the EURNN parameterizes the unitary recurrent weight matrix by a product of unitary matrices. Specifically, the product consists of a unitary diagonal matrix and complex Givens rotation matrices. Unlike the restricted-capacity uRNN, the EURNN parameterization has the capacity to represent all possible unitary matrices but requires a long product of matrices.

\section{Scaled Cayley Unitary RNN (scuRNN)}
\label{scuRNN}

\subsection{Scaled Cayley Transform}

Unlike other architectures that use a long product of simple matrices to parameterize the unitary recurrent weight matrix, the scuRNN architecture maintains a strictly unitary recurrent weight matrix by incorporating the following result.
\begin{theorem}[\citeauthor{kahan06,odorney14}]
\label{cayley}
	Every unitary matrix $W$ can be expressed as 
	$$ W = (I+A)^{-1}(I-A)D$$ where $A =[a_{ij}]$ is skew-Hermitian with $|a_{ij}| \leq 1$ and $D$ is a unitary diagonal matrix. For an orthogonal $W$, the same result holds with $A$ being skew-symmetric and $D$ a diagonal matrix with entries consisting of $\pm 1$.
\end{theorem}

In scoRNN, the orthogonal matrix is constructed using the orthogonal parameterization in Theorem \ref{cayley}. The scaling matrix $D$ is not known a priori and needs to be determined for each particular task.  Since $D$ consists of discrete valued parameters, it can not be determined by gradient descent.  However, $D$ is essentially defined by the number of negative ones on the diagonal barring a permutation.  Thus the number of negative ones is considered a hyperparameter that must be tuned for optimal results with the additional restriction that the scaling matrix must be fixed during training. 

The scuRNN architecture uses the complex version in Theorem \ref{cayley}. It overcomes the constraints inherent with scoRNN since $D$ in this case has entries of the form $D_{j,j} = e^{i\theta_j}$. This parameterization is differentiable with respect to the continuous $\theta_j$ variables and can be determined by gradient descent during training with $D$ no longer being fixed. 

\subsection{Architecture Details}

The scuRNN architecture is similar to a standard RNN, see (\ref{eq:1}), with the exception that the hidden bias is incorporated in the modReLU activation function $\sigma_{\text{modReLU}}(z)$, see section \ref{modReLU} for definition, and all matrices are complex valued.
\begin{equation}
\label{scuRNN}
\textbf{h}_t  =  \sigma_{\text{modReLU}}\left(U\textbf{x}_t + W\textbf{h}_{t-1}\right) \; ; \;\textbf{y}_t  = V\textbf{h}_t + \textbf{c}
\end{equation}

Since the input to hidden, recurrent, and hidden to output weight matrices are complex valued, we follow the framework described in \cite{Arjo16} to compute complex matrix vector products by separating all complex numbers in terms of their real and imaginary parts.

\subsection{Training the Skew-Hermitian and Scaling Matrices}

In order to train the skew-Hermitian matrix $A$ and scaling matrix $D$ that are used to parameterize the unitary recurrent weight matrix in scuRNN, we have to deal with complex derivatives. When we consider the loss function as a function of the complex matrix A or scaling matrix $D$ with a range on the real-line, the loss function is nonholomorphic and thus not complex differentiable.   To compute the necessary gradients, Wirtinger calculus is required \cite{Kreutzdelgado09}.

In Wirtinger calculus, differentiable complex functions are viewed as differentiable functions over $\mathbb{R}^2$.  In particular, given a nonholomorphic function, $f: \mathbb{C} \to \mathbb{R}$, the differential $df$ is given by
\[
	df = \frac{\partial f}{\partial z}dz + \frac{\partial f}{\partial \overline{z}}d\overline{z},
\]

\noindent where $z := x + \it{i}y \in \mathbb{C}$ and $\overline{z} := x - \it{i}y \in \mathbb{C}$ is the conjugate. Here the Wirtinger derivatives are given by  
\[
	\frac{\partial f}{\partial z} = \frac{1}{2}\left(\frac{\partial f}{\partial x} - \it{i}\frac{\partial f}{\partial y} \right) \quad \text{and} \quad \frac{\partial f}{\partial \overline{z}} = \frac{1}{2}\left(\frac{\partial f}{\partial x} + \it{i}\frac{\partial f}{\partial y}\right).
\]

Results from \cite{Hunger07} show that the steepest descent direction using Wirtinger calculus is $\frac{\partial f(z)}{\partial \overline{z}}$. 

Using the above Wirtinger derivatives and steepest descent direction, we update the unitary recurrent weight matrix $W$ by performing gradient descent on the associated skew-Hermitian parameterization matrix $A$ and scaling matrix $D$. In order to compute gradients with respect to $A$, we must pass the gradients through the scaled Cayley transform. The desired gradients for $A$ and diagonal arguments of $D$ are given in Theorem \ref{dldatheorem}. A proof is given in the Appendix.

\begin{theorem}
\label{dldatheorem}
Let $L = L(W): \mathbb{C}^{n \times n}\rightarrow \mathbb{R}$ be a differentiable cost function for an RNN with recurrent weight matrix $W$. Let $W = W(A,\boldsymbol{\theta}) := (I+A)^{-1}(I-A)D$ where $A \in \mathbb{C}^{n \times n}$ is skew-Hermitian, $\boldsymbol{\theta} = [\theta_1, \theta_2, ..., \theta_n]^T \in \mathbb{R}^n$, and $D = \text{diag}\left(e^{i\theta_1}, e^{i\theta_2}, ... , e^{i\theta_n} \right) \in \mathbb{C}^{n \times n}$ is a unitary diagonal matrix. Then the gradient of $L = L(W(A,\boldsymbol{\theta}))$ with respect to $\overline{A}$ is
$$ \frac{\partial L}{\partial \overline{A}} = C^T -\overline{C} $$
where $C:=(I+A)^{-T} \frac{\partial L}{\partial W}(D+W^T)$, $\frac{\partial L}{\partial \overline{A}}=\left[ \frac{\partial L}{\partial \overline{A}_{i,j}} \right] \in \mathbb{C}^{n \times n}$, and $\frac{\partial L}{\partial W}=\left[ \frac{\partial L}{\partial W_{i,j}} \right] \in \mathbb{C}^{n \times n}$.
Furthermore, the gradient of $L = L(W(A,\boldsymbol{\theta}))$ with respect to $\boldsymbol{\theta}$ is given by
\[
	\frac{\partial L}{\partial \boldsymbol{\theta}} = 2\text{Re}\left(\it{i} \left( \left(\frac{\partial L}{\partial W}^{T}K \right)\odot I\right)d \right)
\]
where $K = \left(I + A \right)^{-1}\left(I-A\right)$, $d = \left[e^{\it{i}\theta_1}, e^{\it{i}\theta_2}, ... , e^{\it{i}\theta_n} \right]^T$ is the diagonal vector of $D$ and $\odot$ denotes entry-wise multiplication.
\end{theorem}

We use the above theorem to update the recurrent weight matrix. First we compute $\frac{\partial L}{\partial W},$ using the standard backpropagation algorithm. Then using $\frac{\partial L}{\partial W},$ we compute $\frac{\partial L}{\partial \overline{A}},$ and $\frac{\partial L}{\partial \theta},$ using Theorem \ref{dldatheorem}. We then update the diagonal matrix $D$ by first updating the argument vector $\theta = \left[ \theta_1, ... , \theta_n \right]^{T}$ using a standard optimizer, such as gradient descent, and reforming $D$.
\begin{equation*}
 \boldsymbol{\theta}^{(k+1)} = \boldsymbol{\theta}^{(k)} - \alpha \frac{\partial L(A^{(k)},\boldsymbol{\theta}^{(k)})}{\partial \boldsymbol{\theta}} \; ; \;
D^{(k+1)}=\text{diag}\left(e^{i\boldsymbol{\theta}^{(k+1)}}\right)
\end{equation*}
where $\alpha$ is the learning rate and diag$(\cdot)$ forms a diagonal matrix. We then update the matrix $A$
$$ A^{(k+1)} =A^{(k)} -\beta \frac{\partial L(A^{(k)},\boldsymbol{\theta}^{(k)})}{\partial \overline{A}} ,$$
where $\beta$  is the learning rate.  We should note that for optimizers that involve squaring the entries of the gradient element-wise, such as RMSProp, Adam, and Adagrad, the update of $A$ is split into updating the real component and imaginary component separately to maintain a skew-Hermitian matrix $A$. Since $\frac{\partial L}{\partial \overline{A}}$ is skew-Hermitian and skew-Hermitian matrices are closed under addition, $A^{(k+1)}$ will be skew-Hermitian. Finally, we construct the recurrent weight matrix using the scaled Cayley transform
$$ W^{(k+1)}  = (I+A^{(k+1)})^{-1}(I-A^{(k+1)})D^{(k+1)}.$$

\section{ModReLU activation Function}
\label{modReLU}
The right selection of a nonlinear activation function plays a major role in avoiding the vanishing and exploding gradient problem. We use the modReLU activation function  for a complex variable which is a modification of the ReLU activation function.  The modReLU activation function was first proposed by \cite{Arjo16} and also used in architectures by \cite{Wisdom16,jing16,jing17,Wolter17} to handle complex valued functions and weights and studied in \cite{trabelsi18}. The modReLU activation function is defined as 
\begin{align}
\label{original_modrelu1}
\sigma_{\text{modReLU}}(z) &=\begin{cases} 
      (|z|+b)\frac{z}{|z|} & \text{if } |z|+b \geq 0 \\
      0 & \text{if } |z|+b < 0
   \end{cases} \\
   \label{original_modrelu2}
   &= \frac{z}{|z|}\sigma_{\text{ReLU}}(|z|+b),
\end{align}
where $b$ denotes a trainable bias. If $b>0$, the modReLU activation function as defined above has a discontinuity at $z=0$ no matter how $\sigma_{\text{modReLU}}(0)$ is defined. As a result, the derivative of modReLU has a singularity at $z=0$ when $b>0$ as follows:

\begin{theorem}
\label{dsigmadz}
For the modified rectified linear activation function $\sigma_{\text{modReLU}}(z),$ the Wirtinger derivatives are given by
\begin{align*}
\frac{\partial \sigma_{\text{modReLU}}(z)}{\partial z} &=\begin{cases} 
      1 + \frac{b}{2|z|} & \text{if } |z|+b \geq 0 \\
      0 & \text{if } |z|+b < 0,
   \end{cases} \\
\frac{\partial \sigma_{\text{modReLU}}(z)}{\partial \overline{z}}  &= \begin{cases} 
      \frac{1}{2}\left[ \frac{-bz^2}{|z|^3} \right] & \text{if } |z|+b \geq 0 \\
      0 & \text{if } |z|+b < 0.
   \end{cases}
\end{align*}
In particular, if $b >0$, then $\frac{\partial \sigma_{\text{modReLU}}(z)}{\partial z}$ and $\frac{\partial \sigma_{\text{modReLU}}(\textbf{z})}{\partial \overline{z}}$ tend to infinity as $z \mapsto 0.$
\end{theorem}

A proof of the theorem is given in the Appendix.

\begin{remark}
If b is positive and $|z| \ll b,$ then $|\frac{\partial \sigma_{\text{modReLU}}(z)}{\partial \overline{z}}|$ is extremely large and will result in floating point exceptions such as NAN during training. On the other hand, if $b \le 0$ then the derivatives are well defined and bounded for all $z$.
\end{remark}

\begin{figure}[h]
    \centering
    \subfloat{{\includegraphics[width=7cm]{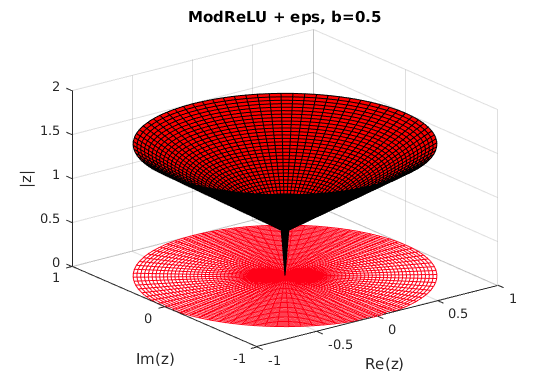} }}%
    \qquad
    \subfloat{{\includegraphics[width=7cm]{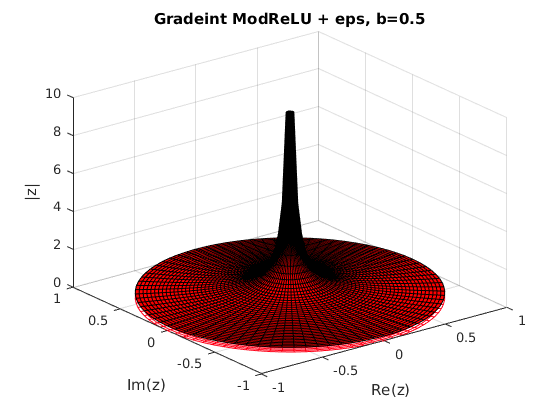} }}%
    \caption{Surface plots of the modulus of the approximate modReLU activation function $\sigma_{\epsilon}$ (top) and the modulus of the gradient of $\sigma_{\epsilon}$ with respect to $\bar{z}$ (bottom). Both plots use a bias of b=0.5.}%
    \label{fig_modrelu1}%
\end{figure}

In the implementations of the uRNNs by \cite{Arjo16} and \cite{Wisdom16}, the following approximate modReLU activation function is used: 
\begin{equation}
\label{implementedmodrelu}
\sigma_{\epsilon}(z)= \frac{z}{\hat{z}+\epsilon}\, \sigma_{\text{ReLU}}(\hat{z}+b) 
\end{equation}
\noindent where $\epsilon = 10^{-5}$ and $\hat{z} := \sqrt{x^2+y^2 + \epsilon}$.  The idea behind this is to avoid division by zero during the forward pass and backward pass of the network. This version of the modReLU function is also implemented in the scuRNN model. Unfortunately, a large derivative can still occur if $\hat{z}+\epsilon \ll b$ as shown in the following theorem and Figure \ref{fig_modrelu1}. 

\begin{theorem}
\label{dsigmadzwithepsilon}
The Wirtinger derivatives of the approximate modReLU activation function can be given as:
  \begin{align*}
  \frac{\partial \sigma_{\epsilon}(z)}{\partial z} &=\begin{cases} 
        \frac{\hat{z}+b}{\hat{z}+\epsilon} + \frac{|z|^2(\epsilon-b)}{2\hat{z}(\hat{z}+\epsilon)^2}& \text{if } \hat{z}+b \geq 0 \\
        0 & \text{if } \hat{z}+b < 0,
     \end{cases} \\
  \frac{\partial \sigma_{\epsilon}(z)}{\partial \overline{z}}  &= \begin{cases} 
        \frac{z^2(\epsilon-b)}{2\hat{z}(\hat{z}+\epsilon)^2} & \text{if } \hat{z}+b \geq 0 \\
        0 & \text{if } \hat{z}+b < 0.
     \end{cases}
  \end{align*}
\end{theorem}

In particular, we found that the unitary RNNs are sensitive to the initialization of the bias, $\textbf{b}$, and the initial state, $\textbf{h}_{0}$. For example, in the MNIST experiment, see section \ref{unpermutemnist}, when the initial state $\textbf{h}_{0}$ is set to  $\textbf{h}_{0} = \textbf{0}$ and non-trainable and \textbf{b} is initialized by sampling from the uniform distribution $\mathcal{U}[-0.01,0.01]$, the gradient values of the loss function would become NAN before the end of the first training epoch. With a random initialization of $\textbf{b}$, many entries are positive and are much larger in magnitude then the corresponding entries of $\hat{\textbf{z}}$ which results in large derivatives. These large derivatives over many time steps can lead to the exploding gradient problem. The cause for small $\hat{\textbf{z}}$ in this experiment is due to the fact that the first several pixels of any given MNIST image will most likely have zero pixel value which combined with the zero initial state  $\textbf{h}_{0}$ will result in small values of $\hat{\textbf{z}}$ compared to the corresponding entries in $\textbf{b}$. 

To avoid small values of $\hat{\textbf{z}}$, it is useful to initialize $\textbf{h}_0$ away from zero. This will mitigate the effects when the first several sequence entries are zero. Based on experimentation, it is also advantageous to allow $\textbf{h}_0$ to be trainable. If singularity is still encountered with initializing and/or training $\textbf{h}_0$, constraining $b$ to be nonpositive may be used, since the singularity only occurs when $b>0$. For example, initializing $\textbf{b}=\textbf{0}$ will avoid the singularity  at least initially so that training can proceed, regardless of the magnitude of $\hat{\textbf{z}}$. However, subsequent training would typically turn some entries of $\textbf{b}$ into positive numbers. On the other hand, we have experimented with schemes that maintain nonpositive $\textbf{b}$, which indeed eliminate the singularity but tend to hinder performace.    

\section{Experiments}
\label{experiments}
In this section, we compare the performances between the restricted-capacity uRNN, full-capacity uRNN, EURNN, LSTM, scoRNN, oRNN and scuRNN architectures on a variety of tasks. Code for these experiments is available at \url{https://github.com/Gayan225/scuRNN}. For each model, the hidden size was adjusted to match the number of trainable parameters. For scuRNN, the real component of the skew-Hermitian matrix was initialized as a skew-symmetric matrix using the initialization used in \cite{kyle17} while the imaginary component was initialized to zero.  The initial hidden state was initialized using the distribution $\mathcal{U}[-0.01,0.01]$ and was trainable. The input to hidden matrix $U$ and hidden to output matrix $V$ were initialized using Glorot \cite{Glorot10}. The $\boldsymbol{\theta}$ values are sampled from $\mathcal{U}\left[ 0,2\pi \right]$, which results in the diagonal entries $D_{j,j}=e^{\it{i}\theta_j}$ being uniformly distributed on the complex unit circle. The biases are initialized from the distribution $\mathcal{U}\left[ -0.01,0.01 \right]$.


The parameterization used in scuRNN allows the use of different optimizers and different learning rates for the input and output weights, skew-Hermitian matrix, and the scaling matrix. We used several different combinations of optimizers and learning rates as noted under each experiment. The reasoning behind mixing different optimizers and learning rates is that the A and D matrices are implicit parameters that are not weights themselves and their entries may have different scales from those of the weights. An update on A and D using the same optimizer/learning rates as the non-recurrent weights may result in an update of W that is incompatible with the updates in the non-recurrent weights. However, scuRNN may be implemented with the same optimizer and
learning rate for A and D matrices, which would involve no additional hyperparameters for tuning compared to scoRNN. In most cases, they produce competitive results ( see Appendix).

Experiment Settings for scoRNN and LSTM are in accordance with \cite{kyle17}, \cite{Wisdom16}, and their corresponding codes. When not listed in their papers, we used the following settings with results that are consistent with the other papers. For LSTM we used an RMSProp optimizer with learning rate $10^{-3}$ on MNIST, permuted MNIST, and copying problems with forget gate bias initialize to 1.0. For the adding problem, we used an Adam optimizer with learning rate $10^{-2}$.  For TIMIT, an RMSProp optimizer with learning rate $10^{-3}$ with forget gate bias -4 were used.  For scoRNN, the copying and adding problems used an RMSProp optimizer with learning rate $10^{-4}$ for A and an Adam optimizer with learning rate $10^{-3}$ for all other weights for the adding problem and an RMSProp optimizer with learning rate $10^{-3}$ for all other weights for the copying problem.  

\subsection{MNIST Classification}
\label{unpermutemnist}
\begin{figure}[h]
\centering
\includegraphics[width=0.8\linewidth]{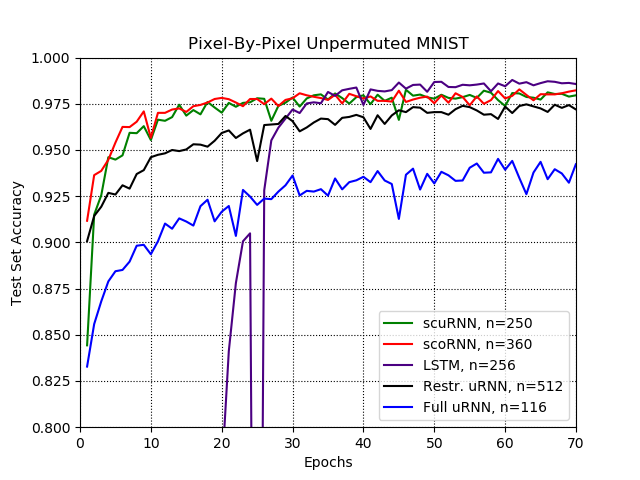}
  \caption{Pixel-by-Pixel MNIST Results}
  \label{fig_mnist1}
\end{figure}

\begin{figure}[h]
\centering
\includegraphics[width=0.8\linewidth]{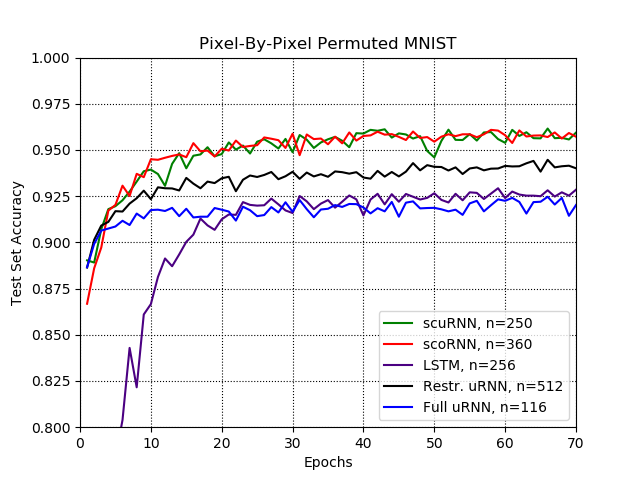}
  \caption{Permuted Pixel-by-Pixel MNIST Results}
  \label{fig_mnist2}
\end{figure}

This experiment involves the classification of handwritten digits using the MNIST database \cite{LeCun}. The data set consists of 55,000 training images and 10,000 testing images with each image in the dataset consisting of a $28 \times 28$ pixel gray-scale image of a handwritten digit ranging from 0 to 9.  Using the procedure outlined in \cite{Le15}, each image is flattened into a vector of length 784 with a single pixel sequentially fed into the RNN. The last sequence output is used to classify the digit. We refer to this experiment as the unpermuted MNIST experiment. A variation of this experiment is to apply a fixed permutation to both the training and test sequences and we refer to this version as the permuted MNIST experiment. All the models were trained for a total of 70 epochs in accordance with \cite{kyle17}. Results of the experiments are given in Table \ref{table:t1}, Figure \ref{fig_mnist1}, and Figure \ref{fig_mnist2}. 

\begin{table}[h]
\begin{center}
\caption{Results for the MNIST classification problem. The best epoch test accuracy over the entire 70 epoch run are recorded. Entries marked by an asterix are reported results from \cite{Mhammadi16} and \cite{jing16}.}
\label{table:t1}
\renewcommand{\arraystretch}{1.2}
\begin{tabular}{ | c | c | c | c | c |} 
\hline
\multicolumn{1}{|c|}{}  & 
\multicolumn{1}{|c|}{} & 
\multicolumn{1}{|c|}{\#}& 
\multicolumn{1}{|c|}{Unperm.}&
\multicolumn{1}{|c|}{Perm.}\\
\multicolumn{1}{|c|}{Model}&
\multicolumn{1}{|c|}{n}&
\multicolumn{1}{|c|}{params}&
\multicolumn{1}{|c|}{Test Acc.}&
\multicolumn{1}{|c|}{Test Acc.}
\\ \hline
scuRNN  
& \multicolumn{1}{|c|}{116} 
& \multicolumn{1}{|c|}{$\approx$ 16k}
& \multicolumn{1}{|c|}{0.976}
& \multicolumn{1}{|c|}{0.949} 
\\
scuRNN  
& \multicolumn{1}{|c|}{250} 
& \multicolumn{1}{|c|}{$\approx$ 69k}
& \multicolumn{1}{|c|}{0.983}
& \multicolumn{1}{|c|}{0.962} 

\\\hline
scoRNN 
& \multicolumn{1}{|c|}{170} 
& \multicolumn{1}{|c|}{$\approx$ 16k}
& \multicolumn{1}{|c|}{0.973} 
& \multicolumn{1}{|c|}{0.943} 

\\
scoRNN 
& \multicolumn{1}{|c|}{360} 
& \multicolumn{1}{|c|}{$\approx$ 69k}
& \multicolumn{1}{|c|}{0.983} 
& \multicolumn{1}{|c|}{0.962} 
\\\hline
LSTM  
& \multicolumn{1}{|c|}{128} 
& \multicolumn{1}{|c|}{$\approx$ 68k}
& \multicolumn{1}{|c|}{0.987} 
& \multicolumn{1}{|c|}{0.920} 
\\
LSTM  
& \multicolumn{1}{|c|}{256} 
& \multicolumn{1}{|c|}{$\approx$ 270k}
& \multicolumn{1}{|c|}{0.989} 
& \multicolumn{1}{|c|}{0.929} 

\\\hline
Rest. cap. uRNN  
& \multicolumn{1}{|c|}{512} 
& \multicolumn{1}{|c|}{$\approx$ 16k}
& \multicolumn{1}{|c|}{0.976} 
& \multicolumn{1}{|c|}{0.945} 

\\\hline
Full. cap. uRNN  
& \multicolumn{1}{|c|}{116} 
& \multicolumn{1}{|c|}{$\approx$ 16k}
& \multicolumn{1}{|c|}{0.947} 
& \multicolumn{1}{|c|}{0.925} 
\\
Full. cap. uRNN  
& \multicolumn{1}{|c|}{512} 
& \multicolumn{1}{|c|}{$\approx$ 270k}
& \multicolumn{1}{|c|}{0.974} 
& \multicolumn{1}{|c|}{0.947} 
\\\hline
oRNN 
& \multicolumn{1}{|c|}{256} 
& \multicolumn{1}{|c|}{$\approx$ 11k}
& \multicolumn{1}{|c|}{0.972*} 
& \multicolumn{1}{|c|}{-} 
\\\hline
EURNN
& \multicolumn{1}{|c|}{512} 
& \multicolumn{1}{|c|}{$\approx$ 9k}
& \multicolumn{1}{|c|}{-} 
& \multicolumn{1}{|c|}{0.937*} 
\\\hline
\end{tabular}
\end{center}
\end{table}

For the unpermuted MNIST experiment, the scuRNN models used an RMSProp optimizer to update the skew-Hermitian matrix and an Adagrad optimizer to update the scaling matrix with all other parameters updated using the Adam optimizer. For hidden size $n=116$, the learning rates were $10^{-4}$, $10^{-3}$,  and $10^{-3}$ respectively. For hidden size $n=250$, the learning rates were $10^{-5}$, $10^{-4}$, and $10^{-3}$ respectively.   Although scuRNN was unable to outperform the LSTM architecture, the $n=116$ scuRNN was able to match the performance of the $n=512$ restricted-capacity uRNN and to outperform all other models.  It should be noted that the scuRNN had a much smaller hidden size with less than 4 times the hidden size of the restricted-capacity uRNN. The $n=250$ scuRNN was able to match the accuracy of the $n=360$ scoRNN with a smaller hidden size. 

For the permuted MNIST experiment, the optimizers for the scuRNN models were the same as the ones used in the unpermuted MNIST experiment. For hidden size $n=116$, the learning rates for the scuRNN model were $10^{-4}$, $10^{-3}$, and $10^{-3}$ respectively. For hidden size $n=250$, the learning rates were the same except the skew-Hermitian matrix had a learning rate of $10^{-5}$.  In this task the $n=250$ scuRNN matches the highest test accuracy of the $n=360$ scoRNN which is higher than all other unitary RNN and LSTM models. It should be noted that the smaller $n=116$ scuRNN outperforms the $n=170$ scoRNN with the same order of trainable parameters.

\subsection{Copying Problem}

The experiment follows the setup described in \cite{Arjo16}, \cite{Wisdom16}, and \cite{kyle17}.  A sequence is passed into the RNN using the digits 0-9. The first ten entries are uniformly sampled from the digits 1-8. This is followed by a sequence of T zeros and a marker digit 9. At the marker digit, the RNN is to output the first ten entries of the sequence.  This results in an entire sequence length of T+20.  The baseline for this task is a network that outputs all zeros except for the last 10 digits which are uniformly sampled from the digits 1-8 for an expected categorical cross entropy of $\frac{10 \log (8)}{T+20}.$

For our experiment we adjust the number of hidden units of each network so that they all have approximately 22k trainable parameters. This results in an LSTM with hidden size $n = 68,$ a restricted-capacity uRNN with $n=470,$ a full-capacity uRNN with $n= 128,$ a EURNN with $n=512$ and capacity $L=2$, a scoRNN with $n=190,$ and a scuRNN with $n=130$. We also tested the architectures using a sequence length of $T=1000$ and $T=2000$ zeros.

The T=1000 scuRNN used Adagrad optimizer with learning rate $10^{-4}$ for the skew-Hermitian matrix and for the diagonal scaling matrix, and Adam optimizer with learning rate $10^{-3}$ for all other weights. For the T=2000 task, scuRNN used Adam optimizer with learning rate $10^{-4}$ for the diagonal scaling matrix and RMSProp optimizer with learning rate $10^{-4}$ and $10^{-3}$ for the skew-Hermitian matrix and all other trainable weights respectively. For the other models, we used the settings detailed in \cite{kyle17}. 

\begin{figure}[h]
    \centering
    \subfloat{{\includegraphics[width=7cm]{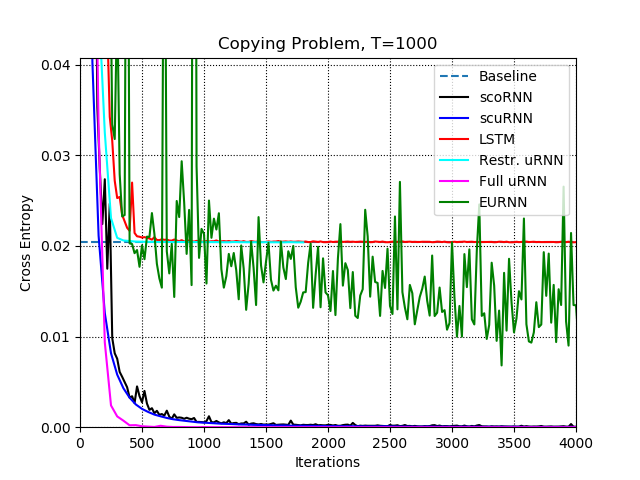} }}%
    \qquad
    \subfloat{{\includegraphics[width=7cm]{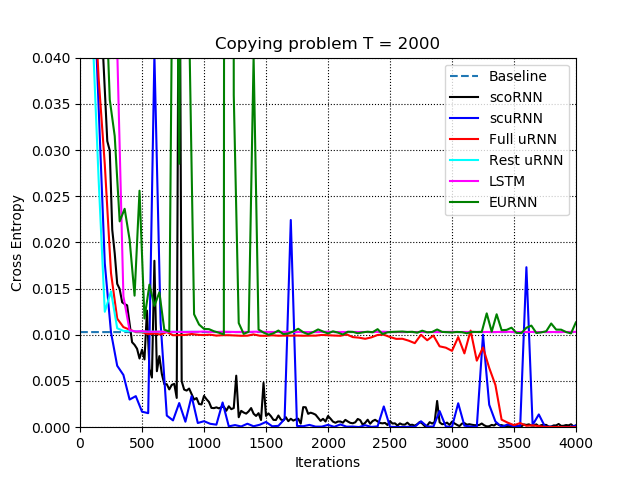} }}%
    \caption{Cross entropy of the copying problem with T = 1000 and T=2000.}%
    \label{fig_copying}%
\end{figure}

The results for this experiment are included in Figure \ref{fig_copying}.  For each experiment, the restricted-capacity uRNN and LSTM converge rapidly to the base line but fail to drop below it. For T =1000, the scuRNN, full-capacity uRNN, and scoRNN quickly drop towards zero cross entropy while the EURNN drops below the baseline but does not converge towards zero. For T=2000, scuRNN drops towards zero cross entropy before all other networks. In this case, the EURNN is unable to drop below the baseline. 

\subsection{Adding Problem}

\noindent The adding problem was first proposed by \cite{Hochreiter97}. In this experiment, we implement a slightly modified version as outlined in \cite{kyle17}. In this problem, two sequences of length $T$ are simultaneously fed into the RNN. The first sequence consists of entries sampled from $\mathcal{U}\left[0,1\right)$. The second sequence consists of all zeros except for two entries that are equal to one. The first one is uniformly located in the first half of the sequence, within the interval $\left[1, \frac T 2 \right)$, while the second one is located uniformly in the other half of the sequence, within the interval $\left[ \frac T 2 , T \right)$. The goal of the network is to compute the sum of the two numbers from the first sequence that are marked by one in the second sequence. The loss function used for this task is the Mean Square Error (MSE). The baseline for this task is an expected MSE of 0.167 which is the expected MSE for a model that always outputs one.

\begin{figure}[h]
    \centering
    \subfloat{{\includegraphics[width=7cm]{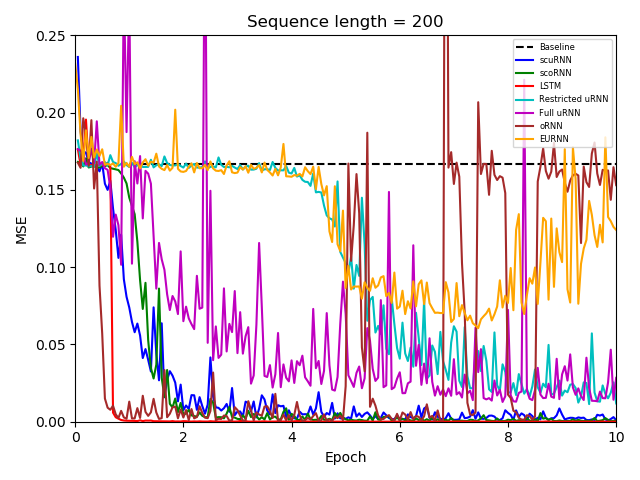} }}%
    \qquad
    \subfloat{{\includegraphics[width=7cm]{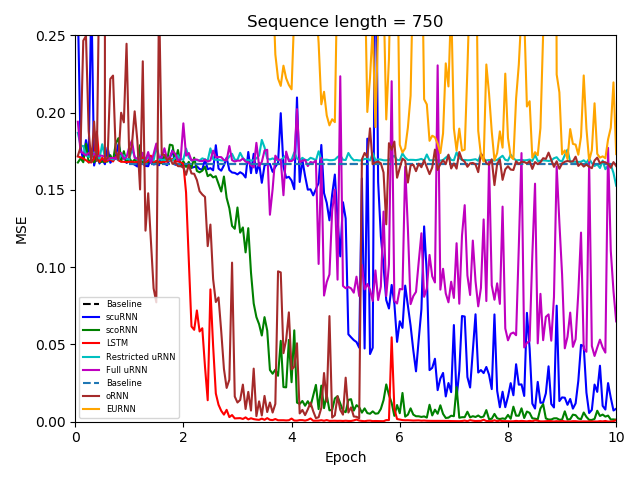} }}%
    \caption{Test set MSE of each machine on the adding problem with sequence length T=200 (above) and sequence length T=750 (bottom).}%
    \label{fig_adding}%
\end{figure}

The sequence lengths used were $T=200$ and $T=750$ with a training set size of 100,000 and a testing set size of 10,000 as in \cite{kyle17}. For T=200, scuRNN used the RMSProp optimizer with learning rate $10^{-3}$ for the skew-Hermitian matrix and the Adam optimizer with learning rate $10^{-3}$ for the diagonal scaling matrix and all other weights. For $T=750$, scuRNN used the Adam optimizer with learning rate $10^{-3}$ for the diagonal scaling matrix and the RMSProp optimizer for the skew-Hermitian matrix and all other weights with learning rates $10^{-4}$ and $10^{-3}$ respectively.  For each model, the number of trainable parameters were matched to be approximately 14k. This results in a hidden size of $n = 116$ for scuRNN, $n = 170$ for scoRNN, $n= 60$ for LSTM, $n = 120$ for the full-capacity uRNN, and $n=950$ for the restricted-capacity uRNN. For the EURNN, the tunable style model with a hidden size of $n=512$ was used for $T=200$ which results in $\approx 3$k trainable parameters and the FFT style was used for $T=750$ which results in $\approx 7$k trainable parameters as outlined in \cite{kyle17}. For oRNN, a hidden size of $n=128$ with 16 reflections which results in $\approx 2.6$k trainable parameters was used in accordance with \cite{kyle17} and \cite{Mhammadi16}.  Results are shown in Figure \ref{fig_adding}.

For sequence length $T=200$, all the architectures start at or near the base line and eventually drop towards zero MSE with the exception of the EURNN model which appears to decrease below the baseline and eventually increases back towards it. The oRNN abruptly drops below the baseline first, followed by scuRNN. Although oRNN is the first to drop below the baseline, the descent curve is erratic with the oRNN bouncing back towards the baseline several training steps later. The LSTM also has a drastic drop towards the zero MSE solution but this occurs after the scuRNN curve passes below the baseline. Although the scuRNN and scoRNN architectures have similar performance, the scuRNN model descends below the baseline before the scoRNN model. For sequence length $T=750$, the oRNN again drops below the base line first but has an erratic descent curve with the oRNN staying at the baseline near the end of training. The LSTM drops below the baseline next followed by the scoRNN and scuRNN models. It should be noted that although the scoRNN model curve descends below the scuRNN curve around the third epoch, the scuRNN model was able to descend towards zero MSE before the full-capacity uRNN and the restricted-capacity uRNN. The EURNN model appears to not be able to decrease below the baseline and the restricted-capacity uRNN model appears to not descend until around the tenth epoch.


\subsection{TIMIT Speech Dataset}

Experiments were conducted using the TIMIT data set \cite{garafolo93}. We use the same setup as described by \cite{Wisdom16} and \cite{kyle17}. TIMIT input values are the log magnitude of the modulus of STFT data and are real-valued. The core test set was used, consisting of a training set of 3,696 audio files,  a testing set of 192 audio files, and a validation set of 400 audio files. Audio files were down sampled as detailed in \cite{Wisdom16}. The hidden size of each model tested were adjusted to match the number of trainable parameters of approximately 83k and 200k.  The best performance for scuRNN was achieved using the Adam optimizer for all trainable parameters. For both $n=128$ and $n=258$ the learning rates were $10^{-4}$ for the skew-Hermitian matrix and $10^{-3}$ for all other parameters.

The loss function for this experiment was the mean square error (MSE) was computed by taking the squared difference between the predicted and
actual log magnitudes and applying a mask to zero out the padded entries then computing the resulting average over the entire batch including zeros.  Table \ref{table:t2} includes the MSE for the validation and evaluation data sets for each model. As can be seen, the scuRNN model outperforms all other models.We
suspect that scuRNN performs well on this data set because the
complex architecture of scuRNN may be better suited to capture the complex dynamics of the underlying STFT sequences. To the best of our knowledge, this is a state of the art result on this task.

\begin{table}[h]
\begin{center}
\caption{Results for TIMIT speech set. Evaluation based on MSE}
\label{table:t2}
\resizebox{.95\columnwidth}{!}{
\begin{tabular}{ | c | c | c | c |c |} 
\hline
\multicolumn{1}{|c|}{Model}  & 
\multicolumn{1}{|c|}{n} & 
\multicolumn{1}{|c|}{\# PARAMS}& 
\multicolumn{1}{|c|}{VALID MSE} &
\multicolumn{1}{|c|}{EVAL. MSE}
\\ \hline
scuRNN  
& \multicolumn{1}{|c|}{128} 
& \multicolumn{1}{|c|}{$\approx$ 83k}
& \multicolumn{1}{|c|}{3.94} 
& \multicolumn{1}{|c|}{3.56} 
\\
scuRNN  
& \multicolumn{1}{|c|}{258} 
& \multicolumn{1}{|c|}{$\approx$ 200k}
& \multicolumn{1}{|c|}{1.84} 
& \multicolumn{1}{|c|}{1.67} 
\\\hline
scoRNN  
& \multicolumn{1}{|c|}{224} 
& \multicolumn{1}{|c|}{$\approx$ 83k}
& \multicolumn{1}{|c|}{4.76}
& \multicolumn{1}{|c|}{4.31}
\\
scoRNN 
& \multicolumn{1}{|c|}{425} 
& \multicolumn{1}{|c|}{$\approx$ 200k}
& \multicolumn{1}{|c|}{2.64}
& \multicolumn{1}{|c|}{2.39}
\\\hline
LSTM  
& \multicolumn{1}{|c|}{84} 
& \multicolumn{1}{|c|}{$\approx$ 83k}
& \multicolumn{1}{|c|}{11.89} 
& \multicolumn{1}{|c|}{10.86}
\\
LSTM  
& \multicolumn{1}{|c|}{158} 
& \multicolumn{1}{|c|}{$\approx$ 200k}
& \multicolumn{1}{|c|}{9.73} 
& \multicolumn{1}{|c|}{8.86}
\\\hline
R.uRNN  
& \multicolumn{1}{|c|}{158} 
& \multicolumn{1}{|c|}{$\approx$ 83k}
& \multicolumn{1}{|c|}{15.57} 
& \multicolumn{1}{|c|}{18.51}
\\
R.uRNN  
& \multicolumn{1}{|c|}{378} 
& \multicolumn{1}{|c|}{$\approx$ 200k}
& \multicolumn{1}{|c|}{16.00} 
& \multicolumn{1}{|c|}{15.15}
\\\hline
F.uRNN  
& \multicolumn{1}{|c|}{128} 
& \multicolumn{1}{|c|}{$\approx$ 83k}
& \multicolumn{1}{|c|}{15.07}
& \multicolumn{1}{|c|}{14.58}
\\
F.uRNN  
& \multicolumn{1}{|c|}{256} 
& \multicolumn{1}{|c|}{$\approx$ 200k}
& \multicolumn{1}{|c|}{14.96} 
& \multicolumn{1}{|c|}{14.69}
\\\hline
\end{tabular}}
\end{center}
\end{table}

\section{Conclusion}

Orthogonal/unitary RNNs have shown promise in mitigating the well-known exploding or vanishing gradient problem. We have developed a new RNN architecture, scuRNN, that is designed to maintain a strict unitary recurrent weight matrix. A simple update scheme  is used to optimize parameters using gradient descent or a related optimizer. This allows us to overcome the inherent difficulty of the scoRNN architecture by removing the need for an extra hyperparameter and allowing the diagonal scaling matrix parameters to be trainable and not fixed during the training process. Our experiments show that scuRNN can achieve comparable or better results than scoRNN and other orthogonal and unitary RNN architectures. We have also discussed a potential singularity in the modReLU activation function, which may have implications on other complex neural networks involving the  modReLU  function.
\subsubsection{Acknowledgments.}
This research was supported in part by NSF under grants  DMS-1821144 and
DMS-1620082. We would also like to thank Devin Willmott for his help on this project.


\section*{Appendix}
In this appendix, we provide detailed proofs of Theorem \ref{dldatheorem} and Theorem \ref{dsigmadz} and additional experimental results for the MNIST classification and copying problems. 
\subsection*{A.1 Proof of Theorem \ref{dldatheorem}}
\begin{proof}
We compute the partial derivative $\frac{\partial L}{\partial A}$ and use the property $\frac{\partial L}{\partial \overline{A}}= \overline{\frac{\partial L}{\partial A}}$ to obtain the desired result. Since $K := (I+A)^{-1}(I-A)$, we have $W = KD$. We consider the $(i,j)$ entry of $\frac{\partial L}{\partial A}.$ Taking the derivative with respect to $A_{i,j},$ where $i \neq j$ we obtain:

\begin{align}
& \frac{\partial L}{\partial A_{i,j}} = \sum_{k,l = 1}^{n} \left(\frac{\partial L}{\partial W_{k, l}}
\frac{\partial W_{k, l}}{\partial A_{i,j}} 
+ \frac{\partial L}{\partial \overline{W}_{k,l}} \frac{\partial \overline{W}_{k, l}}{\partial A_{i,j}}\right) \nonumber \\
 &= \sum_{k = 1}^{n} \left(\frac{\partial L}{\partial W_{k, j}}
D_{l,l}\frac{\partial K_{k, l}}{\partial A_{i,j}}\right) 
+ \sum_{k = 1}^{n} \left( \frac{\partial L}{\partial \overline{W}_{k,l}} \overline{D_{j,j}}\frac{\partial \overline{K_{k, l}}}{\partial A_{i,j}}\right) \nonumber \\
 &= \it{tr}\left[\left(\frac{\partial L}{\partial W} D\right)^T \frac{\partial K}{\partial A_{i,j}}\right]
+ \it{tr}\left[\left(\overline{\frac{\partial L}{\partial W} D}\right)^T \left(\overline{\frac{\partial K}{\partial \overline{A_{i,j}}}}\right)\right].\label{eq:dlda} 
 \end{align}

Let $A = X +\it{i}Y,$ where $X,Y \in \mathbb{R}^{n \times n}.$ It follows that $(I+A)K = I-A$ can be rewritten as $(I+X+\it{i}Y)K = I-X-\it{i}Y.$ Taking the derivatives with respect to $X_{i,j}$ and $Y_{i,j}$ we have:
\begin{equation}
\label{eq1}
\frac{\partial K}{\partial X_{i,j}} = -(I+A)^{-1}\left( \frac{\partial X}{\partial X_{i,j}} + \frac{\partial X}{\partial X_{i,j}} K \right),
\end{equation}
\begin{equation}
\label{eq2}
\frac{\partial K}{\partial Y_{i,j}} = -\it{i}(I+A)^{-1}\left( \frac{\partial Y}{\partial Y_{i,j}} + \frac{\partial Y}{\partial Y_{i,j}} K \right).
\end{equation}

Now let $E_{i,j}$ denote the matrix whose $(i,j)$ entry is 1 with all other entries being zero. Since $X$ is Skew-symmetric, we have $\frac{\partial X}{\partial X_{i,j}} = E_{i,j}-E_{j,i}.$ Similarly, since $Y$ is symmetric, we have $\frac{\partial Y}{\partial Y_{i,j}} = E_{i,j}+E_{j,i}.$ Combining this with equation (\ref{eq1}) and (\ref{eq2}), we obtain:
\begin{align*}
\frac{\partial K}{\partial A_{i,j}} &= \frac{1}{2}\left( \frac{\partial K}{\partial X_{i,j}} -\it{i} \frac{\partial K}{\partial Y_{i,j}}\right)\\
&= \frac{1}{2}\left[ -(I+A)^{-1} \frac{\partial X}{\partial X_{i,j}} (I+ K)\right]\\
&-\frac{1}{2}\left[\it{i}(-\it{i})(I+A)^{-1} \frac{\partial Y}{\partial Y_{i,j}} (I+ K) \right]\\
&= \frac{1}{2}\left[ -(I+A)^{-1} \left(\frac{\partial X}{\partial X_{i,j}} + \frac{\partial Y}{\partial Y_{i,j}}\right)(I+ K) \right]\\
&= -(I+A)^{-1} E_{i,j}(I+K).
\end{align*}
Here the last line is obtained using the Skew-symmetric property of $X$ and the symmetric property of $Y$. Following the same argument one can obtain:
$$
\frac{\partial \overline{K}}{\partial A_{i,j}} 
= \overline{(I+A)^{-1} E_{j,i}(I+Z)}.
$$
We now examine $\frac{\partial L}{\partial A_{i,j}}$ and compute the first trace term in equation (\ref{eq:dlda}) as follows:
\begin{align*}
&\it{tr}\left[\left(\frac{\partial L}{\partial W} D\right)^T \frac{\partial K}{\partial A_{i,j}}\right]\\
&= -\it{tr}\left[\left(\frac{\partial L}{\partial W} D\right)^T (I+A)^{-1}(E_{i,j}+E_{i,j}K) \right]\\
&= -\it{tr}\left[\left(\frac{\partial L}{\partial W} D\right)^T (I+A)^{-1}E_{i,j}\right]\\
&\quad -\it{tr}\left[\left(\frac{\partial L}{\partial W} D\right)^T (I+A)^{-1}E_{i,j}K\right]\\
&= -\left[\left(\left(\frac{\partial L}{\partial W} D\right)^T (I+A)^{-1}\right)^T\right]_{i,j}\\
&\quad -\left[\left(\left(\frac{\partial L}{\partial W} D\right)^T (I+A)^{-1}\right)^T K^T\right]_{i,j}\\
&=-\left[\left(\left(\frac{\partial L}{\partial W} D\right)^T (I+A)^{-1}\right)^T(I+K^T)\right]_{i,j}\\
&=-\left[ ((I+A)^{-1})^T \frac{\partial L}{\partial W} (D +DK^T)\right]_{i,j}  \\
&= -\left[ \left((I+A)^{-1}\right)^T \frac{\partial L}{\partial W}(D+W^T)\right]_{i,j}.
\end{align*}
A similar approach gives the second trace term in equation (\ref{eq:dlda}) as follows:
 \begin{align*}
 & \it{tr}\left[\overline{\left(\frac{\partial L}{\partial W} D\right)}^T \overline{\frac{\partial K}{\partial \overline{A_{i,j}}}}\right]\\
 &= \it{tr}\left[\overline{\left(\frac{\partial L}{\partial W} D\right)^T} \overline{(I+A)^{-1}}\overline{(E_{j,i}+E_{j,i}K)} \right]\\
 &= \it{tr}\left[\overline{\left(\frac{\partial L}{\partial W} D\right)}^T \overline{(I+A)^{-1}} \overline{E_{j,i}}  \right] \\ 
 & \quad + \it{tr}\left[\overline{\left(\frac{\partial L}{\partial W} D\right)}^T \overline{(I+A)^{-1}} \overline{E_{j,i}K}  \right]\\
 &= \left[\overline{\left(\frac{\partial L}{\partial W} D\right)}^T \overline{(I+A)^{-1}} \right]_{i,j}  
 + \left[\overline{K} \overline{\left(\frac{\partial L}{\partial W} D\right)}^T \overline{(I+A)^{-1}}\right]_{i,j}\\
 &=\left[ (I+\overline{K}) \overline{\left(\frac{\partial L}{\partial W} D\right)}^T \overline{(I+A)^{-1}} \right]_{i,j} \\
 &= \left[ \overline{(D^T+W)} \overline{\frac{\partial L}{\partial W}}^T\overline{(I+A)^{-1}}\right]_{i,j}.
 \end{align*}
 Now by the definition of $C$, we have 
 $$\frac{\partial L}{\partial A} = \overline{C}^T -C. $$
 Therefore,
 $$ \frac{\partial L}{\partial \overline{A}} = \overline{\frac{\partial L}{\partial A}} = C^T -\overline{C}. $$

To compute $\frac{\partial L}{\partial \theta_{j}}$, we take the derivative with respect to $\theta_j$ where $1 \leq j \leq n.$  
\begin{align*}
 \frac{\partial L}{\partial \theta_{j}} &= \sum_{k = 1}^{n} \left(\frac{\partial L}{\partial W_{k, j}}
 \frac{\partial W_{k, j}}{\partial \theta_j} 
 + \frac{\partial L}{\partial \overline{W}_{k,j}} \frac{\partial \overline{W}_{k, j}}{\partial \theta_j}\right) \\
  &= \sum_{k = 1}^{n} \left(\frac{\partial L}{\partial W_{k, j}}
 \frac{\partial K_{k, j}D_{j,j}}{\partial \theta_j} 
 + \frac{\partial L}{\partial \overline{W}_{k,j}} \frac{\partial \overline{K_{k, j}D_{j,j}}}{\partial \theta_j}\right) \\
  &= \sum_{k = 1}^{n} \left(\frac{\partial L}{\partial W_{k, j}}
 K_{k, j}\frac{\partial e^{\it{i}\theta_j}}{\partial \theta_j} 
 + \frac{\partial L}{\partial \overline{W}_{k,j}} \overline{K_{k, j}}\frac{\partial e^{-\it{i}\theta_j}}{\partial \theta_j}\right) \\
  &= \it{i}\sum_{k = 1}^{n} \frac{\partial L}{\partial W_{k, j}}K_{k, j} D_{j,j} -\it{i}\sum_{k = 1}^{n} \frac{\partial L}{\partial \overline{W}_{k,j}}\overline{K}
 _{k,j}\overline{D}_{j,j}
\end{align*}

Since this holds for all $0 \leq j \leq n$ we have
\begin{align*}
 \frac{\partial L}{\partial \boldsymbol{\theta}} &= \it{i} \left( \left(\frac{\partial L}{\partial W}^{T}K \right)\odot I\right)d+ \overline{\it{i} \left( \left(\frac{\partial L}{\partial W}^{T}K \right)\odot I\right)d}\\
 &= 2\text{Re}\left(\it{i} \left( \left(\frac{\partial L}{\partial W}^{T}K \right)\odot I\right)d \right)
\end{align*}
 as desired.
\end{proof}

\subsection*{A.2 Proof of Theorem \ref{dsigmadz}}
\begin{proof}
Observe that the partial derivative of $\sigma_{\text{modReLU}}(z)$ with respect to x and y, where $z=x+\it{i}y$ given by
\begin{align*}
\frac{\partial \sigma_{\text{modReLU}}(z)}{\partial x} &=\begin{cases} 
      1 + b \frac{(y^2 -ixy)}{|z|^3} & \text{if } |z|+b \geq 0 \\
      0 & \text{if } |z|+b < 0,
   \end{cases} \\
\frac{\partial \sigma_{\text{modReLU}}(z)}{\partial y}  &= \begin{cases} 
      i + b \frac{(i x^2 -xy)}{|z|^3} & \text{if } |z|+b \geq 0 \\
      0 & \text{if } |z|+b < 0.
   \end{cases}
\end{align*}
 Now computing Wirtinger derivatives, we get the desired result. 
 
 \end{proof}
\subsection*{A.3 Additional experimental results}
In our experiments, we optimized the skew-Hermitian matrix A and diagonal scaling matrix D by tuning with respect to the RMSProp, Adam, and Adagrad optimizers and the learning rates $10^{-3},10^{-4},$ and $10^{-5}$. In general, most of these settings may exhibit good convergence curves. To illustrate, we present additional convergence results for a few combinations of various optimizers and learning rates  used to update A, D, and all other trainable parameters.  The MNIST classification results are in Table \ref{table:t3} and the copying problem in Table \ref{table:t4}.

\begin{table}[H]
\begin{center}
\caption{Additional convergence results for various optimizers/learning rates for A, D and all the other weights for the MNIST classification problem with n=116 (RMS. = RMSprop,Adg. = Adagrad)}
\label{table:t3}
\renewcommand{\arraystretch}{1.2}
\begin{tabular}{ | c | c | c | c | c |} 
\hline
\multicolumn{1}{|c|}{}  & 
\multicolumn{1}{|c|}{} & 
\multicolumn{1}{|c|}{Other}& 
\multicolumn{1}{|c|}{Unperm.}&
\multicolumn{1}{|c|}{Perm.}\\
\multicolumn{1}{|c|}{A}&
\multicolumn{1}{|c|}{D}&
\multicolumn{1}{|c|}{weights}&
\multicolumn{1}{|c|}{Test Acc.}&
\multicolumn{1}{|c|}{Test Acc.}
\\ \hline
RMS./$10^{-4}$
&\multicolumn{1}{|c|}{Adg./$10^{-3}$}
&\multicolumn{1}{|c|}{Adam/$10^{-3}$} 
&\multicolumn{1}{|c|}{0.976}
& \multicolumn{1}{|c|}{0.949} 
\\\hline
RMS./$10^{-4}$
&\multicolumn{1}{|c|}{RMS./$10^{-3}$}
&\multicolumn{1}{|c|}{RMS./$10^{-3}$}
&\multicolumn{1}{|c|}{0.956}
& \multicolumn{1}{|c|}{0.933} 
\\\hline
RMS./$10^{-3}$
&\multicolumn{1}{|c|}{RMS./$10^{-3}$}
&\multicolumn{1}{|c|}{RMS./$10^{-3}$}
&\multicolumn{1}{|c|}{0.913}
& \multicolumn{1}{|c|}{0.901} 
\\\hline
RMS./$10^{-4}$
&\multicolumn{1}{|c|}{RMS./$10^{-4}$}
&\multicolumn{1}{|c|}{RMS./$10^{-3}$}
&\multicolumn{1}{|c|}{0.976}
& \multicolumn{1}{|c|}{0.939} 
\\\hline
RMS./$10^{-4}$
&\multicolumn{1}{|c|}{RMS./$10^{-3}$}
&\multicolumn{1}{|c|}{RMS./$10^{-4}$}
&\multicolumn{1}{|c|}{0.961}
& \multicolumn{1}{|c|}{0.913} 
\\\hline
RMS./$10^{-4}$
&\multicolumn{1}{|c|}{RMS./$10^{-4}$}
&\multicolumn{1}{|c|}{RMS./$10^{-4}$}
&\multicolumn{1}{|c|}{0.976}
& \multicolumn{1}{|c|}{0.920} 
\\\hline
\end{tabular}
\end{center}
\end{table}

\begin{table}[H]
\begin{center}
\caption{Additional convergence results for various optimizers/learning rates for A, D and all the other weights for the Copying problem $T=2000$ (RMS. = RMSprop, Iter.=first iteration below baseline, It. MSE = MSE at iteration 2000)}
\label{table:t4}
\renewcommand{\arraystretch}{1.2}
\begin{tabular}{ | c | c | c | c | c |} 
\hline
\multicolumn{1}{|c|}{}  & 
\multicolumn{1}{|c|}{} & 
\multicolumn{1}{|c|}{Other}& 
\multicolumn{1}{|c|}{}&
\multicolumn{1}{|c|}{}\\
\multicolumn{1}{|c|}{A}&
\multicolumn{1}{|c|}{D}&
\multicolumn{1}{|c|}{weights}&
\multicolumn{1}{|c|}{Iter.}&
\multicolumn{1}{|c|}{It. MSE}
\\ \hline
RMS./$10^{-4}$
&\multicolumn{1}{|c|}{Adam/$10^{-4}$}
&\multicolumn{1}{|c|}{RMS./$10^{-3}$}
&\multicolumn{1}{|c|}{300}
& \multicolumn{1}{|c|}{2.5E-4} 
\\\hline
RMS./$10^{-4}$
&\multicolumn{1}{|c|}{RMS./$10^{-3}$}
&\multicolumn{1}{|c|}{RMS./$10^{-3}$}
&\multicolumn{1}{|c|}{600}
& \multicolumn{1}{|c|}{7.2E-4} 
\\\hline
RMS./$10^{-4}$
&\multicolumn{1}{|c|}{RMS./$10^{-4}$}
&\multicolumn{1}{|c|}{RMS./$10^{-3}$}
&\multicolumn{1}{|c|}{300}
& \multicolumn{1}{|c|}{1.9E-4} 
\\\hline
RMS./$10^{-4}$
&\multicolumn{1}{|c|}{RMS./$10^{-5}$}
&\multicolumn{1}{|c|}{RMS./$10^{-3}$}
&\multicolumn{1}{|c|}{350}
& \multicolumn{1}{|c|}{1.0E-3} 
\\\hline
RMS./$10^{-4}$
&\multicolumn{1}{|c|}{RMS./$10^{-4}$}
&\multicolumn{1}{|c|}{RMS./$10^{-4}$}
&\multicolumn{1}{|c|}{600}
& \multicolumn{1}{|c|}{7.2E-4} 
\\\hline
\end{tabular}
\end{center}
\end{table}


\end{document}